\documentclass[twoside]{article}

\usepackage[accepted]{aistats2022}
\usepackage[round]{natbib}

\usepackage[utf8]{inputenc} 
\usepackage[T1]{fontenc}    
\usepackage{hyperref}       
\usepackage{url}            
\usepackage{booktabs}       
\usepackage{amsfonts}       
\usepackage{nicefrac}       
\usepackage{microtype}      
\usepackage{xcolor}         
\usepackage{graphicx}
\usepackage{import}

\newcommand{\be}{\begin{equation}}
\newcommand{\ee}{\end{equation}}
\newcommand{\ba}{\begin{align}}
\newcommand{\ea}{\end{align}}
\newcommand{\bea}{\begin{eqnarray}}
\newcommand{\eea}{\end{eqnarray}}

\usepackage{amsmath}

\newcommand{\vw}{{\boldsymbol{w}}}
\newcommand{\vx}{{\boldsymbol{x}}}

\begin{document}

\twocolumn[

\aistatstitle{Two-argument activation functions learn soft XOR operations like cortical neurons}

\aistatsauthor{ Kijung Yoon$^\ast$  \And Emin Orhan \And Juhyun Kim \And Xaq Pitkow$^\ast$}

\aistatsaddress{Department of\\ Electronic Engineering\\ Hanyang University \And  
Center for Data Science\\ NYU \And 
Department of\\ Electronic Engineering\\ Hanyang University \And
Department of Neuroscience\\ Baylor College of Medicine\\ Department of ECE\\ Rice University } ]

\begin{abstract}
Neurons in the brain are complex machines with distinct functional compartments that interact nonlinearly. In contrast, neurons in artificial neural networks abstract away this complexity, typically down to a scalar activation function of a weighted sum of inputs. Here we emulate more biologically realistic neurons by learning canonical activation functions with two input arguments, analogous to basal and apical dendrites. We use a network-in-network architecture where each neuron is modeled as a multilayer perceptron with two inputs and a single output. This inner perceptron is shared by all units in the outer network. Remarkably, the resultant nonlinearities often produce soft XOR functions, consistent with recent experimental observations about interactions between inputs in human cortical neurons. When hyperparameters are optimized, networks with these nonlinearities learn faster and perform better than conventional ReLU nonlinearities with matched parameter counts, and they are more robust to natural and adversarial perturbations.
\end{abstract}

\section{Introduction}
Neurons in the brain are not simply linear filters followed by a half-wave rectification, and exhibit properties like divisive normalization \citep{heeger1992normalization, carandini2012normalization}, coincidence detection \citep{larkum1999new, branco2010dendritic}, and history dependence \citep{barlow1961possible, rieke1999spikes}. Instead of fixed canonical nonlinear activation functions such as \texttt{sigmoid}, \texttt{tanh}, and \texttt{ReLU}, other nonlinearities may be both more realistic and more useful \citep{poirazi2003pyramidal,beniaguev2021single,jones2021might}. We are particularly interested in multivariate nonlinearities like $f(\vw_1^\top \vx, \vw_2^\top \vx, ...)$, where the arguments could correspond to inputs that arise, for example, from multiple distinct pathways such as feedforward, lateral, or feedback connections, or from different dendritic compartments. Such multi-argument nonlinearities could allow one feature to modulate the processing of the others. 

\begin{figure}[t]
	\centerline{\includegraphics[width=0.95\columnwidth]{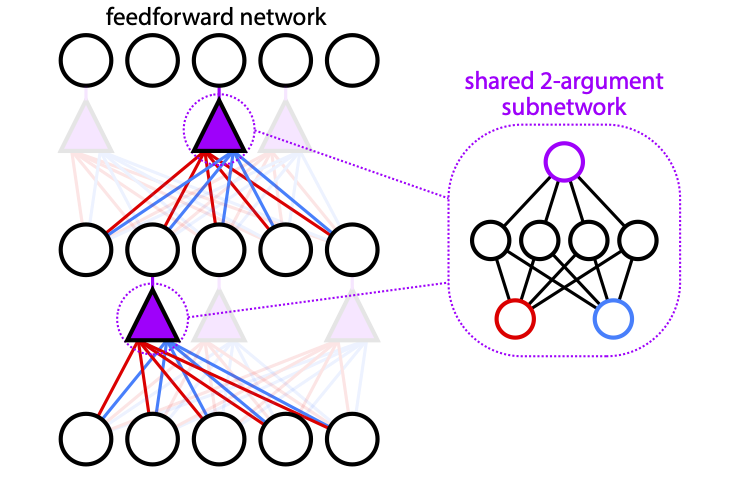}}
	\caption{\textbf{Multi-argument nonlinearities in artificial neurons.} Schematic of architecture including a multi-argument nonlinear activation function (purple triangles). These functions' two arguments are different linear weighted sums of features, and may correspond to distinct inputs such as apical and basal dendrites.}
	\label{fig1}
\end{figure}

\begin{figure*}[t]
    \includegraphics[width=\textwidth]{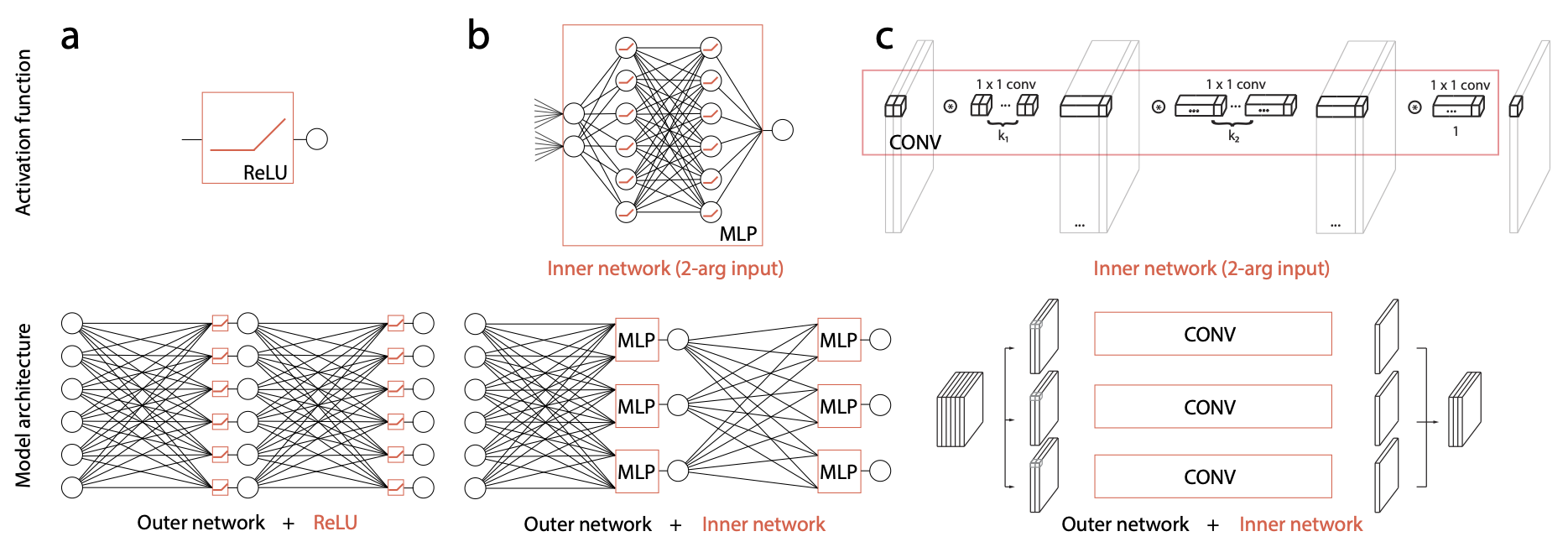}
	\caption{\textbf{Overview of the proposed model structures.} \textbf{(a)} Scalar nonlinear activation function ReLU (top) and MLP-based outer network with ReLU nonlinearities (bottom), \textbf{(b)} $n$-arg input MLP-based inner network (top; $n=2$ in this figure) and the MLP-based outer network that replaces ReLU with the inner network above  (bottom). The activation functions are color-coded by red boxes and the rest of the black other than the red boxes represents the elements of outer network, \textbf{(c)} $1 \times 1$ conv-based inner network (top) merged into conv-based outer network (bottom). The inner network takes inputs from different feature maps; thus the conv-based outer network requires slice and concatenation operations from the depth dimension before and after the inner network. The model schematics assume a two-input argument nonlinearity.}
	\label{fig2}
\end{figure*}

\begin{figure*}[t]
	\includegraphics[width=\textwidth]{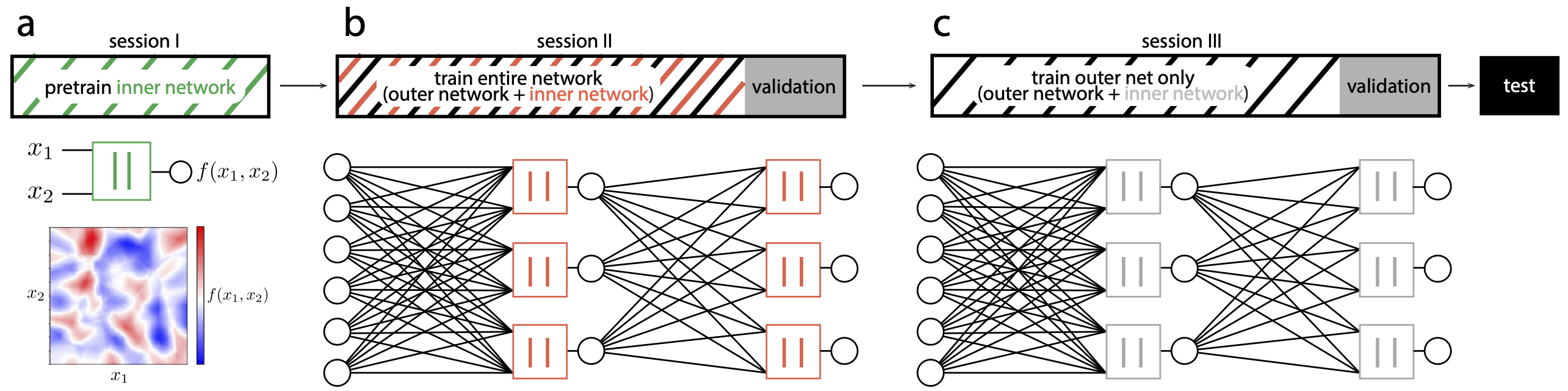}
	\caption{\textbf{Training procedure.} \textbf{(a)} Pretraining. Schematic of two-input argument inner network (green) trained to predict a smoothed random initial activation map (bottom).  \textbf{(b)}  Simultaneously training inner (red) and outer (black) networks. \textbf{(c)} Retraining outer network (black) with frozen inner networks (gray).}
	\label{fig3}
	
\end{figure*}
Recent work showed that a single dendritic compartment of a single neuron can compute the exclusive-or (XOR) operation \citep{gidon2020dendritic}. The fact that an artificial neuron could not compute this basic computational operation discredited neural networks for decades \citep{minsky1969perceptrons}. Although XOR can be computed by networks of neurons, the finding that even single neurons can too highlights the possibility that individual neurons may be much more sophisticated than is often assumed in machine learning. Many single-argument nonlinearities permit universal computation, but the right nonlinearity could allow faster learning and better generalization, both for the brain and for artificial networks.

To investigate this, we parameterize the nonlinear input-output transformation flexibly by an “inner” neural network, which becomes a ‘subroutine’ called from the conventional “outer” network made of many of these complex neurons with parameters that are shared across all layers and all nodes of a given cell type (Figure \ref{fig1}). We evaluate fully-connected and convolutional feedforward networks on image classification tasks given a diverse set of random initial conditions. We focus especially on two-argument nonlinearities learned from MNIST and CIFAR-10 datasets.

\section{Related work}

Numerous recent studies have focused on developing novel activation functions, building on the simplicity and reliability of ReLU \citep{hahnloser2000digital, nair2010rectified}. These studies can be distinguished by the type of learning algorithm used for optimizing the activation function and the size of the search space. Many recent modifications such as PReLU \citep{he2015delving}, ELU \citep{clevert2015fast}, SELU \citep{klambauer2017self}, and GELU \citep{hendrycks2016gaussian} provide single-argument activation functions with a small number of parameters that are mostly fixed (or tuned through hyperparameter optimization). However, such hand-designed functional forms result in restricted expressivity. {\em Swish} \citep{ramachandran2017searching} is noteworthy in this respect, because its activation function is discovered by a combination of exhaustive search and reinforcement learning. The search space in this case is based on a set of predetermined one- and two-argument functions, so this approach can span a broader class of nonlinearities than past work, although it is limited by the specific basis set and the combination rules chosen.

More closely related to our work, the network-in-network architecture proposes to replace groups of simple ReLUs with a fully connected network \citep{lin2013network}. This activation function allows arbitrary dimensional inputs and outputs; thus it is essentially the most general and expressive nonlinear function. However, our work is primarily motivated by neurons in the brain, which can be formalized as multi-input and single-output nonlinear units. As in network-in-network, we parameterize the nonlinear many-to-one transformation by a fully-connected multi-layer network to examine the learned \textit{spatial} activation function without sacrificing its representational power. 

The multi-argument nonlinear transformation is also a canonical operation subsumed under the emerging network architectures such as graph neural networks (GNNs) \citep{scarselli2008graph, li2015gated, kipf2016semi, hamilton2017representation} and transformers \citep{vaswani2017attention, jaegle2021perceiver}. As conceptual extensions from scalar to vector-valued inputs, the message functions in GNNs are multi-input nonlinearities while the scaled dot-product attention in transformers can be viewed as a three-input argument nonlinearity. Although these architectures evaluate performance benefits of specific multi-argument activations, to the best of our knowledge, ours is the first study to characterize the emergent properties of multivariate nonlinear activation functions and their connection to the neuronal nonlinearities in the brain.

\section{Model structure}

To define our multi-argument nonlinearity, we introduce the concepts of inner network and outer network. The inner network aims to learn an arbitrary multivariate nonlinear function $f(x_1, ..., x_n)$ with $n$ inputs and a single output. This will replace the regular scalar activation functions like ReLU. The outer network refers to the rest of the model architecture aside from the activation function. Our framework, composed of two disjoint networks, is flexible and general since diverse neural architectures can be used as outer networks, such as multilayer perceptrons (MLPs), convolutional neural networks (CNNs), ResNets, etc. On the other hand, for the inner network, we use MLPs that have two hidden layers with 64 units followed by ReLU nonlinearities. The MLP is shared across all layers, analogous to the fixed canonical nonlinear activation functions commonly used in feedforward deep neural networks. When we test a CNN-based outer network, we use $1 \times 1$ convolutions instead of MLPs for the inner network to make the model fully convolutional, but the inner network is otherwise essentially the same as the two-layer MLP. In this framework, the $1 \times 1$ conv implies that the inputs to the inner network are channel-wise features, which is similar to the idea of mixing channel information per location in the recent MLP mixer architecture \citep{tolstikhin2021mlp}. Figure \ref{fig2} summarizes how the inner network is incorporated into the outer network.

\section{Experiments}

\subsection{Training procedure}

\begin{figure*}[t!]
	\centerline{\includegraphics[width=0.801\textwidth]{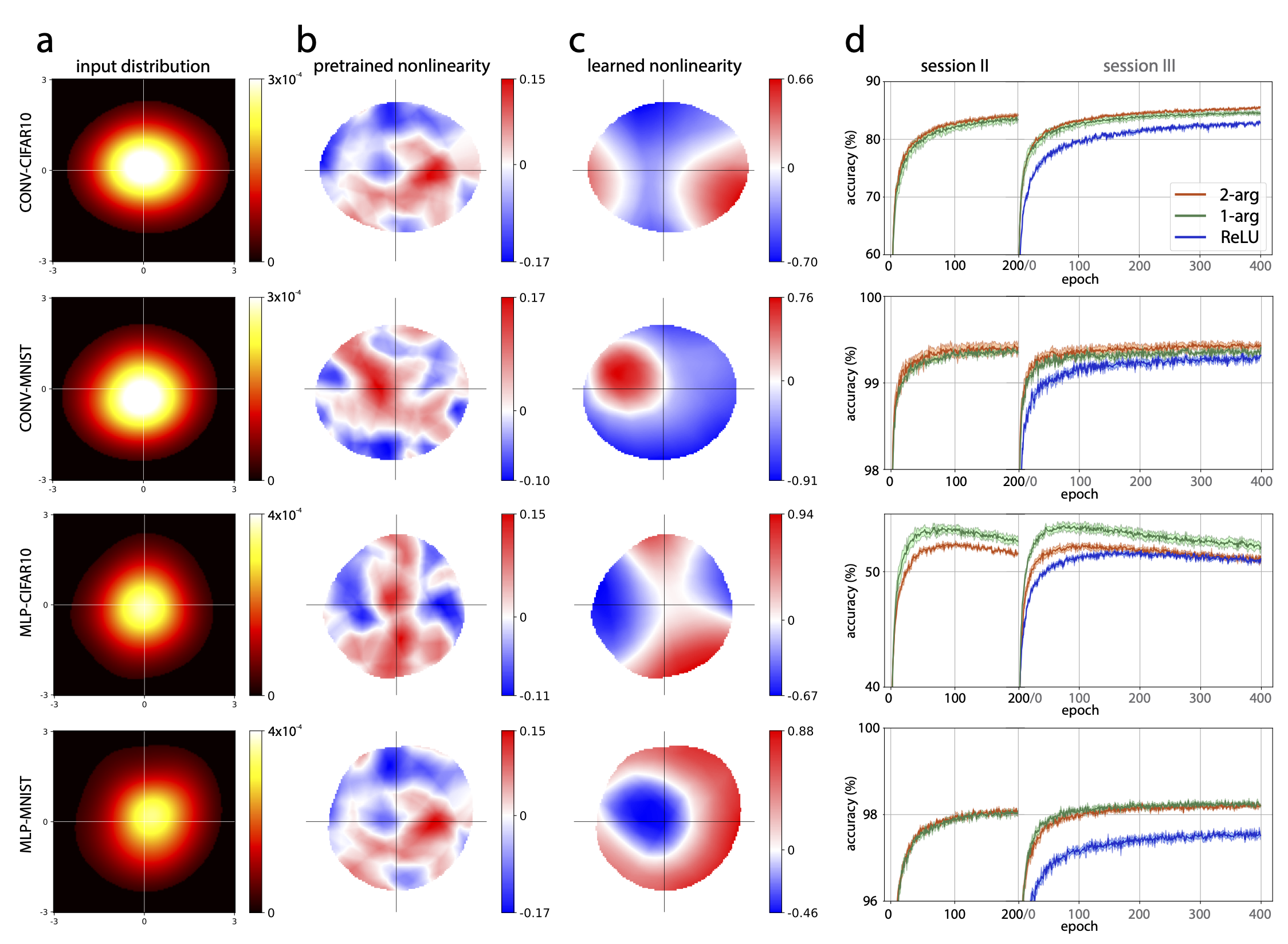}}
	\caption{\textbf{Learned nonlinearities learn tasks faster.} Examples of \textbf{(a)} input distribution, \textbf{(b)} pretrained random initial nonlinearities, and \textbf{(c)} learned two-argument activation functions trained on two different data sets, CIFAR-10 and MNIST, within two different architecture types, a convolutional network and a multi-layer perceptron. Colors indicate the output of activation function, masked to the best-trained part of the input distribution, i.e. for the 99\% of input values that are most common. White bands showing the crossing point between positive (blue) and negative (red) outputs. \textbf{(d)} Average test accuracy (solid line) $\pm 1$ SD (shaded region; $n=4$ samples) of the 2-arg activation model (red) and the baselines (blue: ReLU, green: 1-arg activation) in session \MakeUppercase{\romannumeral 2} (200 epochs) and session \MakeUppercase{\romannumeral 3} (400 epochs). Networks with these two-argument nonlinearities learn faster than others.}
	\label{fig4}
\end{figure*}

\textbf{Pretraining (session \MakeUppercase{\romannumeral 1})} \quad We first generate a random activation function and then use supervised learning to pretrain our inner network to match it (Figure \ref{fig3}a). The motivation for this inner network pretraining stage is that common initialization methods \citep{glorot2010understanding, he2016deep} do not generate spatial activations that are ``random'' enough to study the changes in functions over time. To start with a sufficiently complex initial nonlinearity, we create a piecewise constant random output sampled uniformly from $[-1,1]$ over a $5\times 5$ grid of unit squares tiling the input space. We blur this by a 2D gaussian kernel ($\sigma=3$ units) to define a random smooth activation map. This function serves as the target for the inner network to match (Figure \ref{fig3}a). Example activation functions after pretraining are shown in Figure \ref{fig4}b. This produces our initialized inner network, whose parameters are transferred to the next phase of training.

\textbf{Training inner and outer networks (session \MakeUppercase{\romannumeral 2})} \quad Next we merge the pretrained inner network with outer network via parameter sharing (Figure \ref{fig3}b) and apply this general network-in-network architecture to the task of image classification. In this session, both networks are trained simultaneously so that the entire network is made to learn over what might be analogous to an evolutionary timescale on which nonlinear cell properties emerge (Figure \ref{fig3}b). As our baseline outer networks, we use (1) MLPs that have three hidden layers with 64 units or (2) CNNs that have four convolutional layers with [60, 120, 120, 120] kernels of size $3\times 3$ and a stride of 1, using $2\times 2$ max-pooling with a stride of 2. Aside from the MLPs or convolutional layers, the outer network uses other standard architectural components: layer normalization \citep{ba2016layer} (placed before inner networks) and dropout \citep{srivastava2014dropout} (placed after each hidden/convolutional layer; $p=0.5$). 
Our models are trained on the MNIST and CIFAR-10 datasets using ADAM \citep{kingma2014adam} with a learning rate of 0.001 until the validation error saturates; early-stopping is used with a window size of 20. We freeze the learned nonlinearity $f_{\textrm{inner-net}}(\cdot)$ at the time of saturation or at a maximum epoch. Examples of learned nonlinearities are shown in Figure \ref{fig4}c.

To obtain some intuition about the learned $2$-arg input nonlinearities, we first collect the values of every input to the nonlinearities (i.e. to the inner networks) over all test data at inference time. For display, we compute the pre-activation input distribution (Figure \ref{fig4}a), and show the nonlinearities over the region enclosing 99\% of the input distribution (Figure \ref{fig4}b--c). If two-argument nonlinearities learned what is essentially a one-argument structure, we would see parallel bands of constant color. Instead, notably, all the examples show nontrivial two-dimensional structure, reflecting interactions between the two input arguments (see Section \ref{sec4-3}).

\textbf{Training outer network for fixed inner network (session \MakeUppercase{\romannumeral 3})} \quad Having learned multi-argument nonlinear activation functions, we now fix these inner networks and retrain the outer network to use them on new task data. We borrow the $f_{\textrm{inner-net}}(\cdot)$ from its parameters trained in session \MakeUppercase{\romannumeral 2}, freeze the inner network, and then re-initialize the outer network. In this session, only the outer network is trained as for typical training of a deep neural network with a canonical activation function (Figure \ref{fig3}c). The training curves in this stage are not qualitatively different from what we observed in session II (Figure \ref{fig4}d), indicating that most of the learning over long time intervals (epochs) is attributable to the change of parameters in outer network. In other words, the learning of multi-argument nonlinear activation function may be terminated in an early stage and the rest of learning may be dedicated to solving the classification tasks.

\begin{figure}[t]
	\includegraphics[width=\columnwidth]{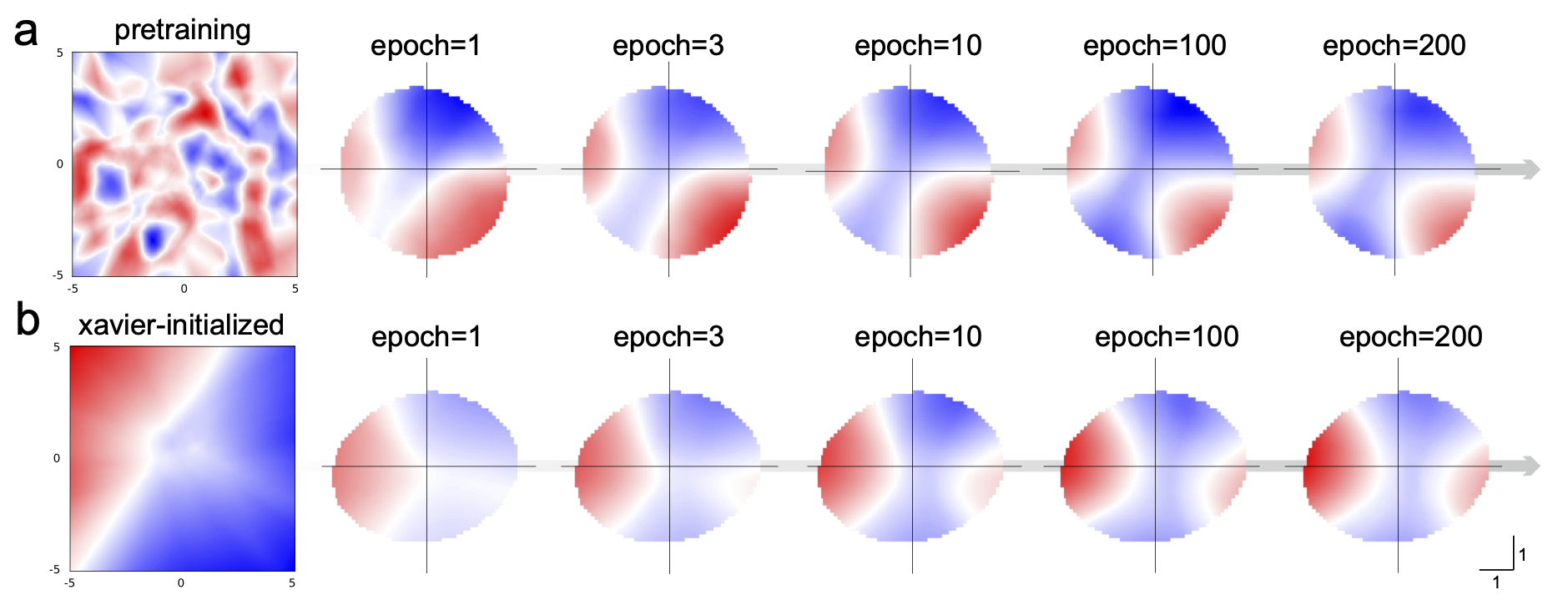}
	\caption{\textbf{Evolution of learned two-argument activation functions.} \textbf{(a)} Snapshot of random initial and learned nonlinear activation functions across development. \textbf{(b)} The same evolution of nonlinearity when it is Xavier-initialized. }
	\label{fig5}
\end{figure}

We thus look for evidence of structural stability of inner network in early development by plotting the learned nonlinearities every epoch in session \MakeUppercase{\romannumeral 2}. We find that the two-argument activation functions mature into typical two-dimensional spatial patterns within 1-5 epoch in general (Figure \ref{fig5}), suggesting that the overall spatial structure of the the activation function emerges quite rapidly from pressures that arise early in the learning process.

\subsection{Comparing to other nonlinearities}
\begin{figure}[t]
	\includegraphics[width=\columnwidth]{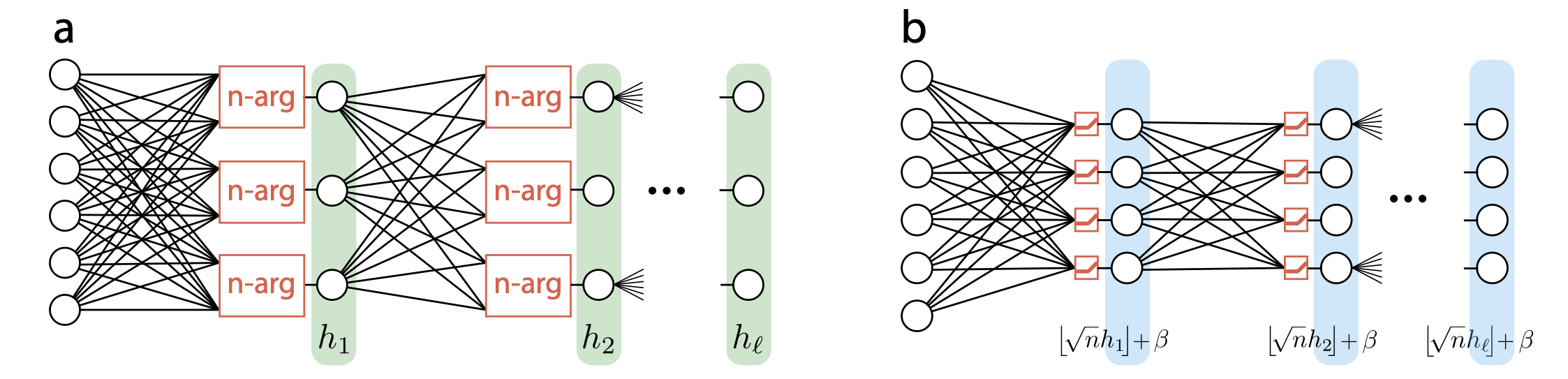}
	\caption{\textbf{Baseline architecture for parameter counts.} \textbf{(a)} MLP-based outer network that have $L$ hidden layers with $h_{\ell}$ units (green) along with $n$-arg input nonlinearities (red). \textbf{(b)} Baseline model architecture with ReLU composed of $L$ hidden layers with $\lfloor \sqrt{n}h_{\ell} \rfloor + \beta$ units (blue) in each layer $\ell$.}
	\label{fig6}
\end{figure}

With the aim of providing context for the performance of our proposed approach we compare against a single-argument nonlinearity. For fair comparison, we train the baseline models, whose architectures are depicted in Figure \ref{fig6}, just as we train our outer networks. The baseline models all involve the same MLP or CNN architecture, i.e. they use the same type and number of outer network layers as our proposed model.

When comparing different architectures we take care to use comparable numbers of learnable parameters in the classification tasks by systematically adjusting the number of hidden units or feature maps in each layer. Specifically, MLP-based outer network with $n$-arg input nonlinearities (Figure \ref{fig5}a) contains $x(nh_1+1)+\sum_{\ell=1}^{L-1} n h_{\ell} h_{\ell+1} + h_Ly + (65n + 4288)$ parameters, where $x, y,$ and $h_{\ell}$ are the dimension of input, output, and the number of units in hidden layer ${\ell}$, respectively. The last term represents the number of inner network parameters; this is independent of input and output dimensions as well as the number of hidden layers $L$, so it does not increase the model complexity (due to parameter sharing). In contrast, the second term $\sum_{\ell=1}^{L-1} n h_{\ell} h_{\ell+1}$ dominates the parameter counts, so our baseline model (Figure \ref{fig6}b) has $L$ layers, each comprising $\lfloor \sqrt{n}h_{\ell} \rfloor + \beta$ hidden units where $\beta$ is a constant to approximate the parameter counts of the proposed model: $\lfloor \sqrt{n}h_{\ell} \rfloor \times \lfloor \sqrt{n}h_{\ell+1} \rfloor \approx n h_{\ell}h_{\ell+1}$. This way of matching parameter counts in MLP-based outer network applies also to CNN-based models, by setting $h_{\ell}$ to be the number of feature maps in convolutional layer $\ell$ instead of hidden units.

Figure \ref{fig4}d compares training performance of the two-input argument nonlinearity to networks using a ReLU or single-argument nonlinearity. We repeat the training of the nonlinearities on MNIST and CIFAR-10 4 times, which produces 4 different samples of model performance. We average the results across 4 samples and find that the models with learned activation functions achieve an overall strong performance (Figure \ref{fig4}d). Notably, Figure \ref{fig4}d suggests that our proposed network learns faster than the ReLU network and achieves better asymptotic performance, providing evidence for a better inductive bias in the network due to the learned multi-argument nonlinearities.

\begin{figure*}[t]
	\includegraphics[width=\textwidth]{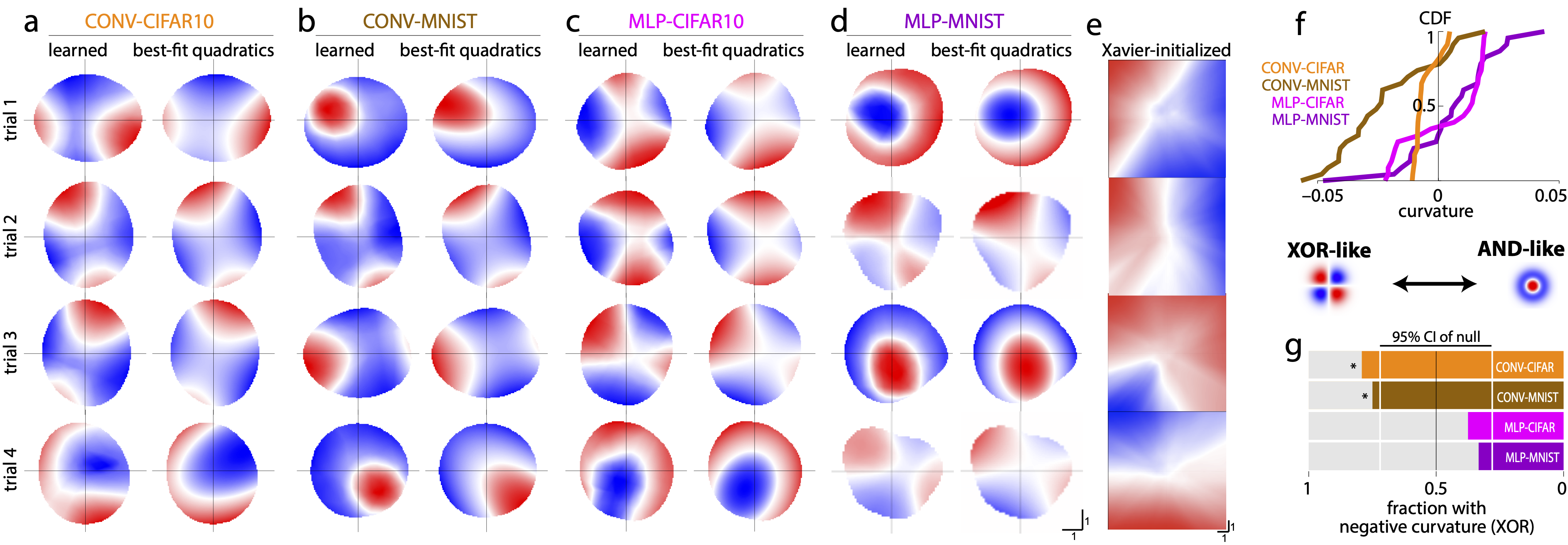}
	\caption{\textbf{Gating operations emerge naturally from learnable multi-argument nonlinear structures.} \textbf{(a-d) Left:} Examples of learned multi-argument activation functions trained on CIFAR-10 and MNIST, within two different architecture types, CNN and MLP. Each row is a different repetition of the learning experiment. All examples show nontrivial two-dimensional structure, reflecting interactions between two input arguments. The majority show a (potentially rotated) white X shape, indicating a multiplicative interaction between the input features, and consistent with a gating interaction or soft XOR. \textbf{(a-d) Right:} The best-fit quadratics of the corresponding left nonlinearities. \textbf{(e)} Random activation functions generated from Xavier weight initialization. \textbf{(f)} Cumulative Distribution Function (CDF) of nonlinearity curvature. \textbf{(g)} Fraction of nonlinearities with negative (XOR-like) curvature. Even a set of random functions may by chance have nonzero average curvature. The CONV architectures show deviations that are outside of the 95\% Confidence Interval (CI) of the null distribution (binomial distribution with probability of 1/2 for positive or negative curvature, for 24 trials).}
	\label{fig7}
\end{figure*}

\subsection{Explicit polynomial nonlinearities}\label{sec4-3}
The results outlined in the previous section focus on the predictive performance of multivariate nonlinear functions. We next turn our attention to the analysis of the structure learned by our multi-argument nonlinearities. We repeat four different trials of the learning experiment and collect samples of two-argument activation functions trained on MNIST and CIFAR-10, within MLP and CNN outer networks. Figure \ref{fig7}a--d (left columns) demonstrates that learned two-argument nonlinearities are reliably shaped like quadratic functions, varying by shifts and/or rotations. We therefore fit an algebraic quadratic functional form, $f(x_1,x_2)=c_1x_1^2+c_2x_2^2+c_3x_1x_2+c_4x_1+c_5x_2+c_6$, to the learned inner-network nonlinearities and find that the learned nonlinearity and its best-fit quadratics have extremely similar structure (Figures \ref{fig7}a--d right). This is the case even though the spatial patterns have different rotations (Figures \ref{fig7}a--d).

We next validate the specificity of the observed inner network output responses. It is clear by eye that the learned nonlinearities are substantially different than those produced by random functions (Figure \ref{fig4}b--c). However, this regular pattern of learned nonlinearities might also be obtainable by popular network initialization methods, such as Xavier weight initialization. To differentiate between these two possibilities, we therefore compare the learned nonlinearities with inner nets initialized with Xavier random initialization \citep{glorot2010understanding} (Figure \ref{fig7}e). We find that the Xaiver random initial activations, although not as ``random'' as those we generated ourselves (Figure \ref{fig4}b), are far from the regular quadratic patterns observed in the learned nonlinearities (Figure \ref{fig7}e). They instead evolve to display such smooth quadratic patterns (Figure \ref{fig5}b), suggesting that the quadratic structures we observe are not captured by standard weight initialization schemes, but are favored by the optimization process instead.

To test whether the learned quadratic functions have statistically significant sub-structure (for example, hyperbolic vs.~elliptical or negative vs.~positive curvature), we computed the curvature implied by the quadratic form above, $c_1c_2-c_3^2/4$ (Figure \ref{fig7}f--g). The convolutional architecture learned nonlinearities with negative curvatures for both tasks, a total of 78\% of 48 trials ($p=0.007$ according to a binomial null distribution with even odds of either curvature). This indicates a multiplicative interaction between the input features, and is consistent with a gating interaction or soft XOR. In contrast, the multilayer perceptron architecture produced more positive curvatures, but these were not statistically significant ($p=0.06$ by the same test).

\subsection{Spectral Analysis}

\begin{figure}[t]
	\includegraphics[width=\columnwidth]{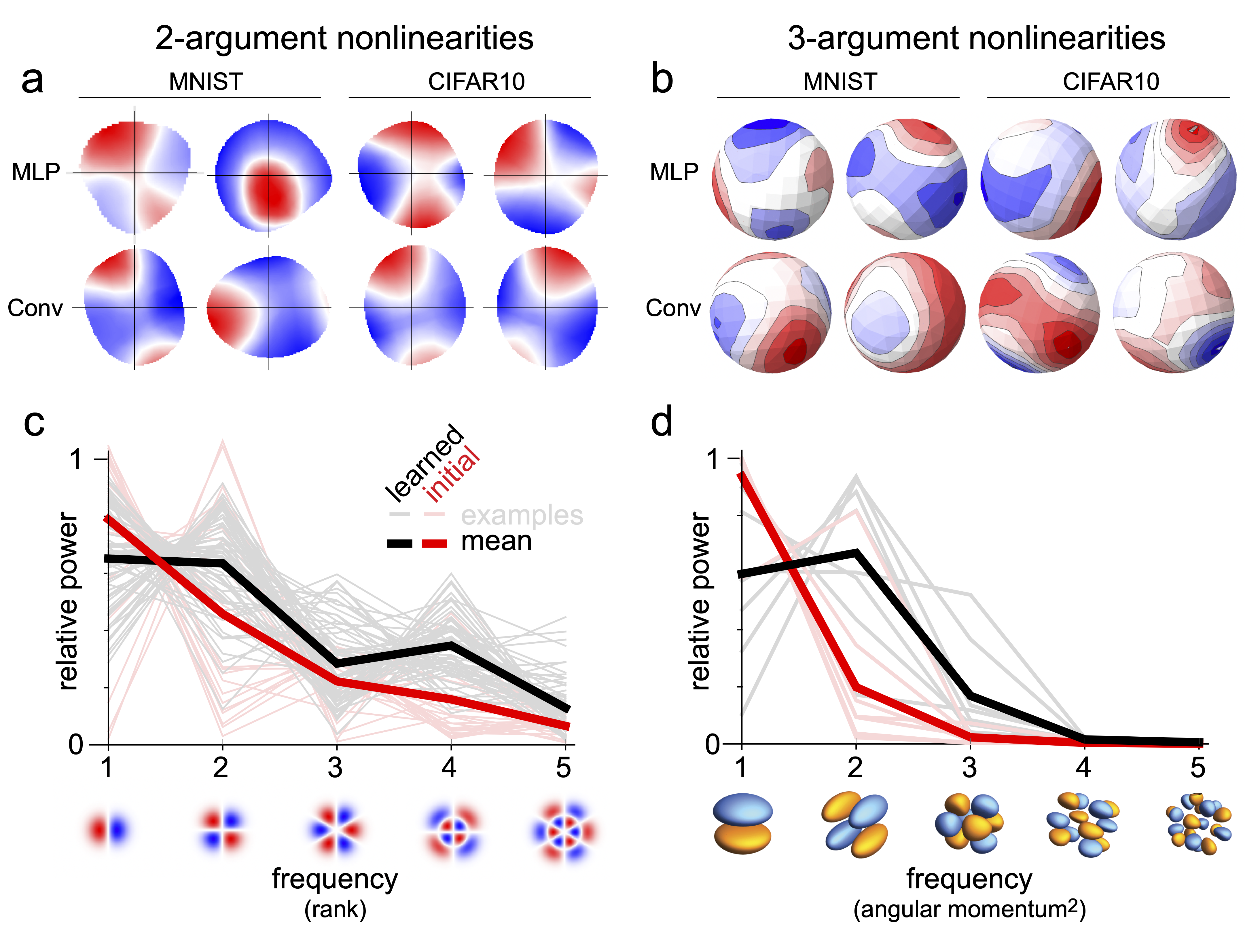}
	\caption{\textbf{Spectral Analysis.} Nonlinearities for various architectures and tasks for \textbf{(a)} two-argument and \textbf{(b)} three-argument inner networks. \textbf{(c--d)} Power spectra for these learned functions (black curves) reveal larger power at $\ell=2$ than spectra for Xavier-initialized inner networks (red), consistent with stronger quadrupolar structure. For the two-argument case, we used 64 learned functions and 24 randomly initialized functions. For the three-argument case, we used 8 learned functions for each. Example basis functions are shown beneath the horizontal axis to illustrate the spatial structure quantified by the frequency number.}
	\label{fig8}
\end{figure}

To further compare the structure of learned nonlinearities to the structure of Xavier-initialized ones, we also performed a spectral analysis on both. We computed spectra using basis functions $\phi(\vx)$ appropriate for the symmetry and boundary conditions of the nonlinearities: we used Hermite-Bessel functions \citep{victor2006responses} for the 2-argument functions, and solid harmonics for the 3-argument functions. We only evaluated the power in regions of the input space that were explored by the distribution $p(\vx)$ of their actual inputs. The power was therefore computed according to $P_\ell[f(\vx)]=\sum_m\|\int d\vx\, p(\vx)f(\vx)\phi_{\ell m}(\vx)\|^2$, where $\ell$ is the analog of spatial frequency for these basis functions and $m$ is analogous to spatial phases. Figure \ref{fig8} shows that the learned multi-argument nonlinearities have more higher-order structure than the Xavier initialized ones. Randomly initialized networks favor strong dipole structure with $\ell=1$. In contrast, the power spectra of learned nonlinearities are consistent with an underlying quadrupole structure, which has its strongest frequency content at $\ell=2$. A soft XOR can be described by $f(x_1,x_2)=x_1 x_2$ or its rotations, which produces positive outputs in two opposite quadrants and therefore creates a quadrupole moment with negative curvature.

\subsection{Generalization}
We now consider out-of-distribution generalization performance of the models for image classification with multi-argument nonlinear functions. In particular, we test whether these activation functions make the learned representations more robust against common image corruptions and adversarial perturbations. We quantify the robustness of the models against common corruptions and perturbations using the recently introduced CIFAR-10-C benchmark \citep{hendrycks2019benchmarking} and parameter-free AutoAttack \citep{croce2020reliable}. 

\begin{figure}[t]
	\includegraphics[width=\columnwidth]{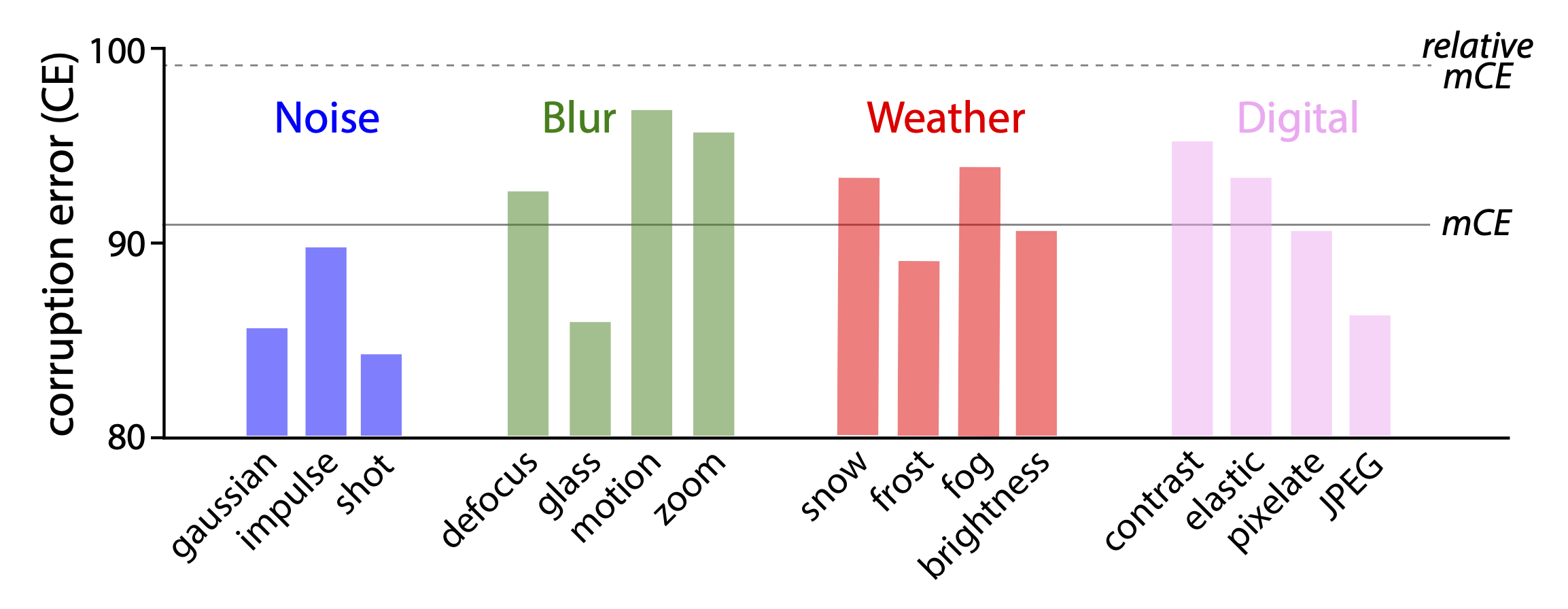}
	\caption{\textbf{Robustness of two-argument nonlinearities against common image corruptions.} Corruption error (CE; bars), mCE (black solid line), and relative mCE (black dashed line) of different corruptions on CIFAR-10-C and Conv-based outer networks. The mCE is the mean corruption error of the corruptions in Noise, Blur, Weather, and Digital categories. Models are trained only on clean CIFAR-10 images.}\vspace{-4mm}
	\label{fig9}
\end{figure}

\textbf{Robustness against common image corruptions}\quad 
CIFAR-10-C was designed to measure the robustness of classifiers against common image corruptions and contains 15 different corruption types applied to each CIFAR-10 validation image at 5 different severity levels. The robustness performance on CIFAR-10-C is measured by the corruption error (CE). For each corruption type $c$, the classification error of the two-argument network is averaged over different severity levels $s$ and then divided by the average classification error of a reference classifier (conv-based outer network with ReLU): i.e. $\mathrm{CE}_c^{2\text{-arg}}=\left.\left(\sum_{s=1}^{5} E_{s, c}^{2\text{-arg}} \right) \middle/\left(\sum_{s=1}^{5} E_{s, c}^{\mathrm{ReLU}}\right)\right.$. The mean corruption error is then obtained by averaging over the corruption types: $\mathrm{mCE}=\langle \mathrm{CE}_c^{2\text{-arg}} \rangle_c$. We also compute a relative mCE score by subtracting the clean classification error of the classifiers from the corruption errors: Relative $\mathrm{CE}_{c}^{2\text{-arg}}=\left.\left(\sum_{s=1}^{5} E_{s, c}^{2\text{-arg}}-E_{\text {clean }}^{2\text{-arg}}\right) \middle/\left(\sum_{s=1}^{5} E_{s, c}^{\text {ReLU}}-E_{\text {clean }}^{\text {ReLU}}\right)\right.$ and then averaging over different corruption types as before results in the $\textrm{relative mCE}=\langle \textrm{relative CE}_c^{2\text{-arg}} \rangle_c$. This measures the relative enhancement on corrupted images in comparison with clean images.

As seen in Figure \ref{fig9}, two-input argument nonlinearities significantly improve the robustness over the ReLU baseline model (mCE $=91.3\%$). Note that mCE scores lower than 100 indicate more success at generalizing to corrupted distribution than the reference model. Moreover, the observed relative mCE ($=99.5\%$, which is less than 100) shows that the accuracy decline of the proposed model in the presence of corruptions is on average less than that of the network with ReLU. The results suggest that this corruption robustness improvements be attributable not only to the simple model accuracy improvements on clean images, but to stronger representations of the learnable multivariate nonlinearity than ReLU against natural corruptions.

\textbf{Adversarial robustness}\quad 
We next consider both black-box and white-box attacks to measure the robustness of the model against adversarial perturbations. We use the recently introduced AutoAttack \citep{croce2020reliable} combining two parameter-free versions of Projected Gradient Descent (PGD) \citep{madry2017towards} algorithm with two existing complementary Fast Adaptive Boundary (FAB) \citep{croce2020minimally} and Square \citep{andriushchenko2020square} attacks. AutoAttack is carried out with an ensemble of the four aforementioned attacks to reliably evaluate adversarial robustness where the hyperparameters of all attacks are fixed for all experiments across datasets and models.

In Table \ref{tab1}, we report the results on 6 models $(\in\{\textrm{MLP, Conv} \}_{\textrm{outer-net}} \times \{\textrm{2-arg, 1-arg, ReLU} \}_{\textrm{inner-net}}) $ trained for $\ell_{\infty}$-robustness. For each classifier we report the accuracy on the robustness test, at the $\epsilon$ specified in the table, on the whole test set obtained by the ensemble AutoAttack. This method counts an attack successful when at least one of the four attacks finds an adversarial example (worst case evaluation). Additionally, we compute the difference in robustness between the network with two-input argument nonlinearities and the baseline model using ReLU nonlinearities. Positive differences are highlighted in blue in the last column of Table \ref{tab1}, and indicate improved robustness compared to the baseline model. In all cases, AutoAttack reveals greater robustness in networks with the learned two-argument nonlinearities than in the baseline networks with ReLU. This suggests that the learned two-argument nonlinearities provide a better inductive bias against adversarial perturbations.

\begin{table}[t]
  \caption{Robustness of adversarial defenses by \textit{AutoAttack}. Numbers indicate average classification accuracy from 4 trials.}
  \label{tab1}
  \resizebox{\columnwidth}{!}{%
  \begin{tabular}{lccccc}
    \toprule
    \multicolumn{2}{c}{}  & \multicolumn{3}{c}{AutoAttack} \\
    \cmidrule(r){3-5}
    Dataset     & Outer-Net & 2-arg & 1-arg & ReLU & increment   \\
    \midrule
    MNIST ($l_{\infty}, \epsilon=0.3$) & MLP  & 39.80 & 22.86 & 26.74 &  \color{blue}{13.06}   \\
    MNIST ($l_{\infty}, \epsilon=0.3$)    & Conv & 49.25 & 10.02 & 9.33  & \color{blue}{39.92}    \\
    CIFAR-10 ($l_{\infty}, \epsilon=0.031$)     & MLP  & 4.83 & 5.62 & 2.96 & \color{blue}{1.87}  \\
    CIFAR-10 ($l_{\infty}, \epsilon=0.031$)    & Conv  & 11.27 & 9.55 & 8.57 & \color{blue}{2.70}  \\
    \bottomrule
  \end{tabular}}
\end{table}

\section{Discussion}

The neurons in biological neural networks are much more intricate machines than the units they inspired in machine learning. Instead,
neural networks in machine learning have been dominated by scalar activation functions. At the same time, it is widely acknowledged that different design choices here can lead to different inductive biases, and architectures with new neural elements are proposed frequently. These elements are usually based on guesses or intuition. Interestingly, one of the most influential elements has been a multiplicative gating nonlinearity, seen in LSTMs \citep{hochreiter1997long}, GRUs \citep{chung2014empirical}, and transformers \citep{vaswani2017attention}. Our experiments demonstrated that gating-like functions emerge automatically from learned multi-argument nonlinear activation functions, as the soft XOR can be interpreted as an output that selects one input dimension of its input and modulates or gates that output by another input dimension. These learned functions have properties resembling dendritic interactions in biological neurons \citep{gidon2020dendritic}. Networks endowed with these functions learn faster and are more robust.

Although these learnable nonlinearities add some complexity to a network, overall these extra inner network parameters are few in number since they are shared across all neurons in the outer network. Moreover, using algebraic polynomial approximations to the learned nonlinearities, as in section~\ref{sec4-3}, can reduce both the number of parameters and the memory requirements of the inner networks in practical applications.

Nontrivial computations in a multilayer network require some sort of nonlinearity, since otherwise the whole network merely performs one linear transformation. The simplest nonlinearity is quadratic, whether the quadratic has negative curvature like a soft XOR, or a positive curvature like coincidence detection. It is interesting that even when allowing for more input arguments, the resultant learned nonlinearities still favor low-order quadratic functions (Figure \ref{fig8}b--d). This could be explained by an implicit bias toward smooth functions \citep{williams2019gradient,sahs2020shallow} while still bending the input space to provide useful computations. Perhaps the learned nonlinearities are as random as possible while fulfilling these minimal conditions. It will be interesting to test this hypothesis by examining the transformations of multiple cell types, or those produced by higher-dimensional functions like network-in-network \cite{lin2013network}, and to see whether different tasks incentivize different computations.

Our study demonstrates that flexible multi-argument activation functions converge to reliable and interpretable patterns and provide computational benefits. However, our study has important limitations that should be addressed in future work. The performance benefits should be evaluated in more architectures and tasks, and at larger scales. There might be synergistic benefits from additional features like skip connections or global modulation. Some of the additional complexity afforded by multi-argument activation functions might be more useful when used in richer architectures, including those with recurrence, dedicated input types (e.g. distinct feedforward, feedback, and lateral interaction arguments), multiple cell types \citep{douglas1991functional,shepherd2004synaptic}, and more intricate dendritic substructures \citep{poirazi2001impact,poirazi2003pyramidal}. Such biologically-inspired additions to neural network architectures could provide inductive biases closer to the inductive biases in biological brains \citep{sinz2019engineering,litwin2019constraining}.

\subsubsection*{Acknowledgements}
Work by XP, KY, and EO on this project was supported in part by NSF CAREER grant 1552868, NSF NeuroNex grant 1707400, NIH BRAIN Initiative Grant 5U01NS094368, an award from the McNair Foundation, the Intelligence Advanced Research Projects Activity (IARPA) via Department of Interior/Interior Business Center (DoI/IBC) contract number D16PC00003, the National Research Foundation of Korea (NRF) grant (No. NRF-2018R1C1B5086404, NRF-2021R1F1A1045390), and the Brain Convergence Research Program of the National Research Foundation (NRF) (No. NRF-2021M3E5D2A01023887) funded by the Korean government (MSIT). The U.S. Government is authorized to reproduce and distribute reprints for Governmental purposes notwithstanding any copyright annotation thereon. Disclaimer: the views and conclusions contained herein are those of the authors and should not be interpreted as necessarily representing the official policies or endorsements, either expressed or implied, of IARPA, DoI/IBC, or the U.S. Government.

\bibliographystyle{apalike}
\bibliography{references}

\end{document}